\documentclass[lettersize,journal]{IEEEtran}

\usepackage{amsmath,amsfonts}
\usepackage{algorithmic}
\usepackage{array}
\usepackage[caption=false,font=normalsize,labelfont=sf,textfont=sf]{subfig}
\usepackage{textcomp}
\usepackage{stfloats}
\usepackage{url}
\usepackage{verbatim}
\usepackage{graphicx}
\def\BibTeX{{\rm B\kern-.05em{\sc i\kern-.025em b}\kern-.08em
    T\kern-.1667em\lower.7ex\hbox{E}\kern-.125emX}}
\usepackage{balance}
\usepackage{algorithm}
\usepackage{booktabs}
\usepackage{cite}
\usepackage{url}
\usepackage{siunitx}

\newcommand{\Tmain}{\mathcal{T}_{\text{main}}}
\newcommand{\Taux}{\mathcal{T}_{\text{aux}}}
\newcommand{\Tmod}{\mathcal{T}_{\text{mod}}}

\begin{document}
\title{
Human-Inspired Framework for Robotic Craniotomy: Integrating Multimodal Fusion and Adaptive Trajectory Adjustment}
\author{Renzhen~Le, %
        Xiao~Zhang, %
        Di~Wu, %
        Yuanyu~Wei, %
        Jiachen~Zhu, %
        Zhenzhi~Ying, %
        Pengfei~Zhang, %
        and~Liming~Shu,~\IEEEmembership{Member,~IEEE}
\thanks{This work has been submitted to the IEEE for possible publication. 
Copyright may be transferred without notice, after which this version may no longer be accessible.

This work was supported by National Natural Science Foundation of China under Grant 52405456 and Grant W2621012. (\textit{corresponding author: Liming~Shu.})

Renzhen~Le, Xiao~Zhang, Di~Wu, Yuanyu~Wei, Jiachen~Zhu and Liming~Shu are with the Dalian University of Technology, Dalian 116024, China (e-mail: \protect\url{renzhenle0218@gmail.com}; \protect\url{xz807113@gmail.com}; \protect\url{diwu@mail.dlut.edu.cn}; \protect\url{yuanyuwei@mail.dlut.edu.cn}; \protect\url{2563072882zjc@mail.dlut.edu.cn}; \protect\url{l.shu@dlut.edu.cn}).

Zhenzhi~Ying and Pengfei~Zhang are with the Department of Mechanical Engineering, The University of Tokyo, Tokyo 113-8656, Japan (e-mail: 
\protect\url{ying@mfg.t.u-tokyo.ac.jp};
\protect\url{zhangpengfei@g.ecc.u-tokyo.ac.jp}).
}}

\markboth{
}%
{Le \MakeLowercase{\textit{et al.}}: Human-Inspired Framework for Robotic Craniotomy}

\maketitle

\begin{abstract}
Manual craniotomy is a high-risk, skill-dependent procedure associated with surgeon fatigue and potential dural injury. While robotic approaches have improved safety, existing open-loop systems rely solely on preoperative images and cannot compensate for intraoperative registration errors or tissue deformation. To address this, we propose a human-inspired closed-loop robotic craniotomy framework that intelligently integrates preoperative planning with intraoperative execution. An adaptive dual-contour fusion algorithm is employed to generate trajectories that conform to complex cranial geometries while maintaining a consistent tool–bone relative pose. For intraoperative perception, a multimodal two-stage cross-modal attention block (CMA)-temporal convolutional network (TCN)-Transformer network combined with an adaptive Bayesian filter fuses force and acoustic signals to achieve robust breakthrough detection under varying bone conditions. Upon detection, an in-situ projection-based trajectory adjustment strategy dynamically compensates for depth deviations, enabling safe residual bone isolation. Experiments on bovine ribs show a breakthrough prediction accuracy of 97\%, a detection latency of 0.048±0.097~s, and a maximum overshoot of 0.29~mm. All four \textit{ex vivo} cranial experiments were successfully completed without dural injury. These results demonstrate that the proposed cybernetic framework enables safe and autonomous craniotomy with highly effective closed-loop control.

\end{abstract}  


\section{Introduction}
\IEEEPARstart{C}{raniotomy} is a fundamental procedure in neurosurgery, and is increasingly required in emerging applications such as brain--computer interface (BCI) implantation, involving the creation of a bone window to expose the dura mater and access intracranial lesions \cite{1}. In conventional procedures, the surgeon first retracts the scalp to expose the skull, then drills several burr holes around the target region, and finally connects these holes using a cranial router to remove the bone flap \cite{2}. However, this two-step manual craniotomy procedure is cumbersome and inefficient, relies heavily on the surgeon’s experience and requires sustained intraoperative attention to avoid excessive penetration, which may cause dural injury or even brain injury \cite{3}. Previous studies have reported unintended durotomy rates of 50\%--70\% in manual craniotomy, which can adversely affect postoperative recovery \cite{4}. To address these issues, robots have been introduced to craniotomy owing to their high precision and stability.

Early robot-assisted craniotomy studies mainly focused on collaborative robotic systems. Cui et al. \cite{5} and Xu et al. \cite{6,7} showed that collaborative robots can help stabilize execution and improve dura protection, while Marcus et al. \cite{8} highlighted the benefits of robotic platforms for optimizing surgical workflow. However, these systems still depend heavily on real-time surgeon intervention and therefore provide only limited autonomy. To overcome this limitation, researchers have further investigated preoperative imaging-based trajectory planning. Cunha-Cruz et al. \cite{9} developed an active robotic milling framework based on preoperative planning, Popovic et al. \cite{10} established a CT-based planning system for skull modeling and trajectory design, and Li et al. \cite{11,12} proposed CT-driven adaptive methods for automatic cutting-path generation and optimization. However, the cranial bone often exhibits non-uniform thickness and complex three-dimensional geometries on both inner and outer surfaces. This requires the cranial router to continuously adjust its pose during craniotomy to accommodate varying depth and curvature. The aforementioned approaches primarily focus on optimizing trajectories on the outer cranial surface while neglecting these geometric constraints, which may increase the risk of unintended dural injury during surgical procedures. 

Although preoperative trajectory planning can guide the robot to autonomously perform craniotomy, approaches that solely rely on open-loop position control are insufficient to handle discrepancies between the planned and actual intraoperative conditions. Such discrepancies may arise from registration errors, structural deformation of the robotic system, and intraoperative skull motion.To mitigate the risks of open-loop execution, closed-loop craniotomy systems utilizing real-time sensor feedback have also been introduced. Bian et al. \cite{13,14} achieved breakthrough detection and vertical force control in robotic craniotomy using force feedback. Xia et al. \cite{15} employed the total harmonic amplitude of the milling sound to monitor cutting depth. Furthermore, they also utilized the fundamental harmonic amplitude of tool-handle vibration signals to fit a polynomial model relating milling depth and angle, thereby enabling coordinated control of both parameters in curved bone milling\cite{16}. In addition, Sun et al. \cite{17} proposed an attention-based CNN-LSTM network for sound-based bone-milling recognition, Qu et al. \cite{18} developed a BP neural network for milling-force perception and bone recognition, and Ying et al. \cite{19} employed an RBF neural network for the binary classification of cutting and breakthrough states. Zakeri et al. \cite{20} further combined drilling sounds with machine learning to distinguish cortical from cancellous bone automatically. Although these methods have enabled sensor-based intraoperative state monitoring and closed-loop control, several challenges still remain. Most existing methods rely on single-signal monitoring, and many validation studies are conducted on flat bone plates or simplified specimens, which fail to fully capture the complex curved geometry of the human skull, thereby limiting robustness in real surgical environments. Furthermore, current systems typically trigger an immediate stop upon breakthrough detection, often leaving a relatively thick residual bone layer that still requires manual removal by the surgeon.

To address challenges in robust breakthrough monitoring, skull-geometry-aware planning, and safe residual bone removal, this article presents a human-inspired robotic craniotomy framework integrating multimodal perception and adaptive trajectory adjustment. The main contributions are summarized as follows:
\begin{itemize}
\item[(1)] A dual-contour-fusion-based adaptive spiral trajectory planning method is proposed for complex skull geometries. By leveraging both outer and inner skull contours, the method generates a main trajectory (\(\Tmain\)) for bulk removal and an auxiliary trajectory (\(\Taux\)) for residual bone removal, while avoiding kinematic singularities via tangential basis decomposition.
\item[(2)] A cross-modal attention (CMA)-temporal convolutional network (TCN)-Transformer network with adaptive Bayesian filtering (ABF) is developed for intraoperative monitoring. By fusing force and sound signals, the method enhances robustness, while ABF suppresses transient disturbances and enables reliable, low-latency breakthrough detection.
\item[(3)] A breakthrough-triggered \textit{in-situ} projection-based trajectory adjustment strategy is proposed to compensate for axial deviations and enable safe residual bone removal without dural injury.
\item[(4)] Experimental validation on bovine ribs and \textit{ex vivo} goat skulls demonstrates accurate monitoring, fast response, and safe craniotomy over complex geometries.
\end{itemize}

The rest of this article is organized as follows. Section II describes the workflow of the task. Section III introduces the system configuration. Section IV details the three core technical methods. Section V presents the experimental results. Section VI provides the discussion. Finally, Section VII concludes this article.

\begin{figure*}[t] 
    \centering
    \includegraphics[width=\textwidth]{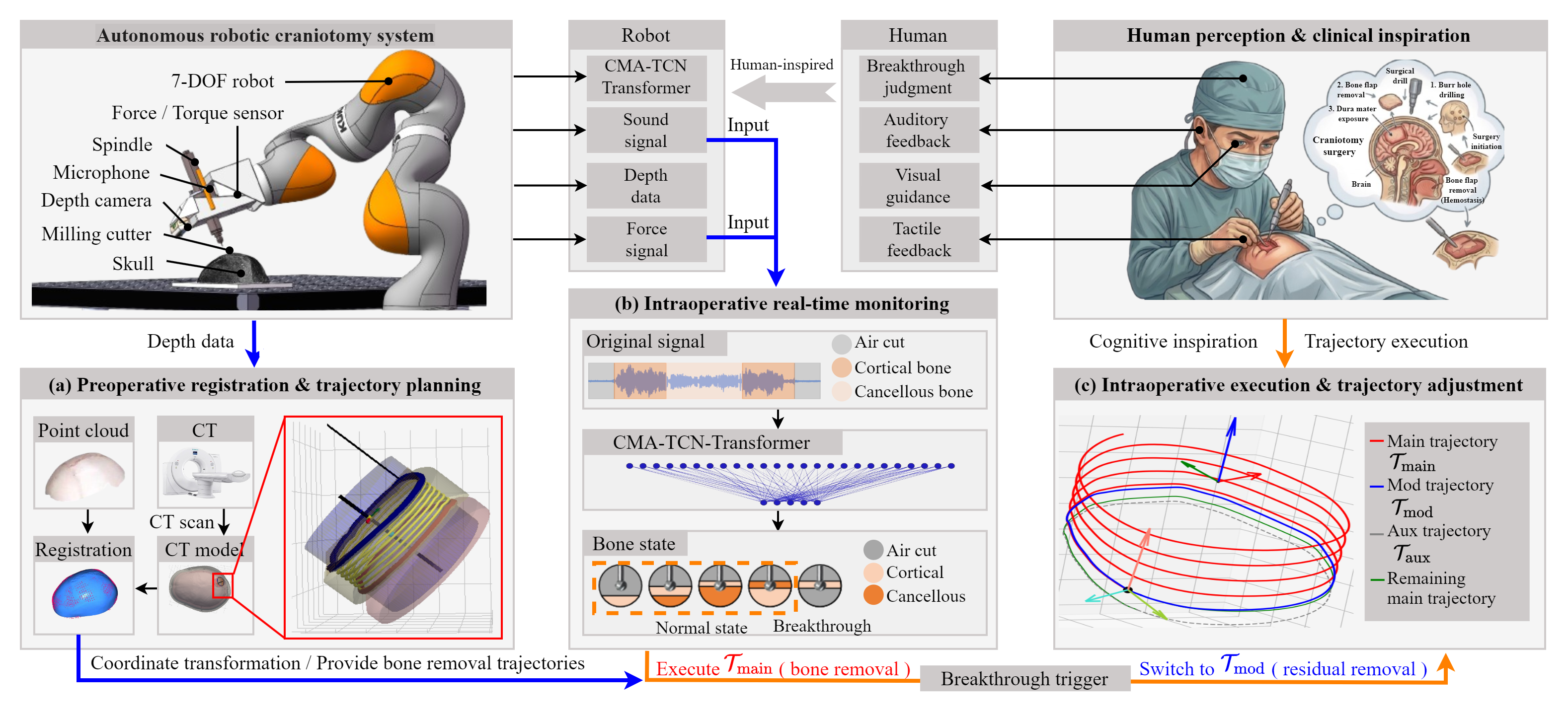}
    \caption{Overall architecture diagram of the autonomous robotic craniotomy system. (a) Preoperative trajectory planning and intraoperative registration. (b)Intraoperative real-time monitoring. (c)Intraoperative execution and trajectory adjustment.}
    \label{fig:system_workflow}
\end{figure*}

\section{Task Description}

Leveraging the high dexterity and positioning accuracy of robotic systems, this paper replaces the conventional multi-step craniotomy procedure with a novel and efficient single-stage approach, where a ball-end milling tool performs spiral milling to achieve direct bone flap removal. Fig.~\ref{fig:system_workflow} illustrates the workflow of the proposed human-inspired robotic craniotomy framework, which consists of three components: preoperative trajectory planning, intraoperative real-time monitoring, and intraoperative execution with trajectory adjustment. 

In the preoperative stage, two functionally complementary trajectories are planned on the patient-specific CT skull model according to the lesion geometry and skull morphology using the proposed dual-contour-fusion-based adaptive spiral trajectory planning method. The main trajectory \(\Tmain\) is designed for efficient bulk bone removal, whereas the auxiliary trajectory \(\Taux\) is reserved for safe residual bone removal near the skull base. In the intraoperative stage, the robot first performs spatial registration and then mills along the predefined main trajectory \(\Tmain\), while the CMA-TCN-Transformer with ABF continuously monitors the current bone-layer state using multimodal sensing signals. Once the breakthrough moment is identified, the predicted breakthrough point is used by the proposed breakthrough-triggered in-situ projection-based trajectory adjustment strategy to generate a modified safe execution path \(\Tmod\), along which the robot completes residual bone removal and achieves automated bone-flap separation.



\begin{figure*}[htbp] 
    \centering
    \includegraphics[width=\textwidth]{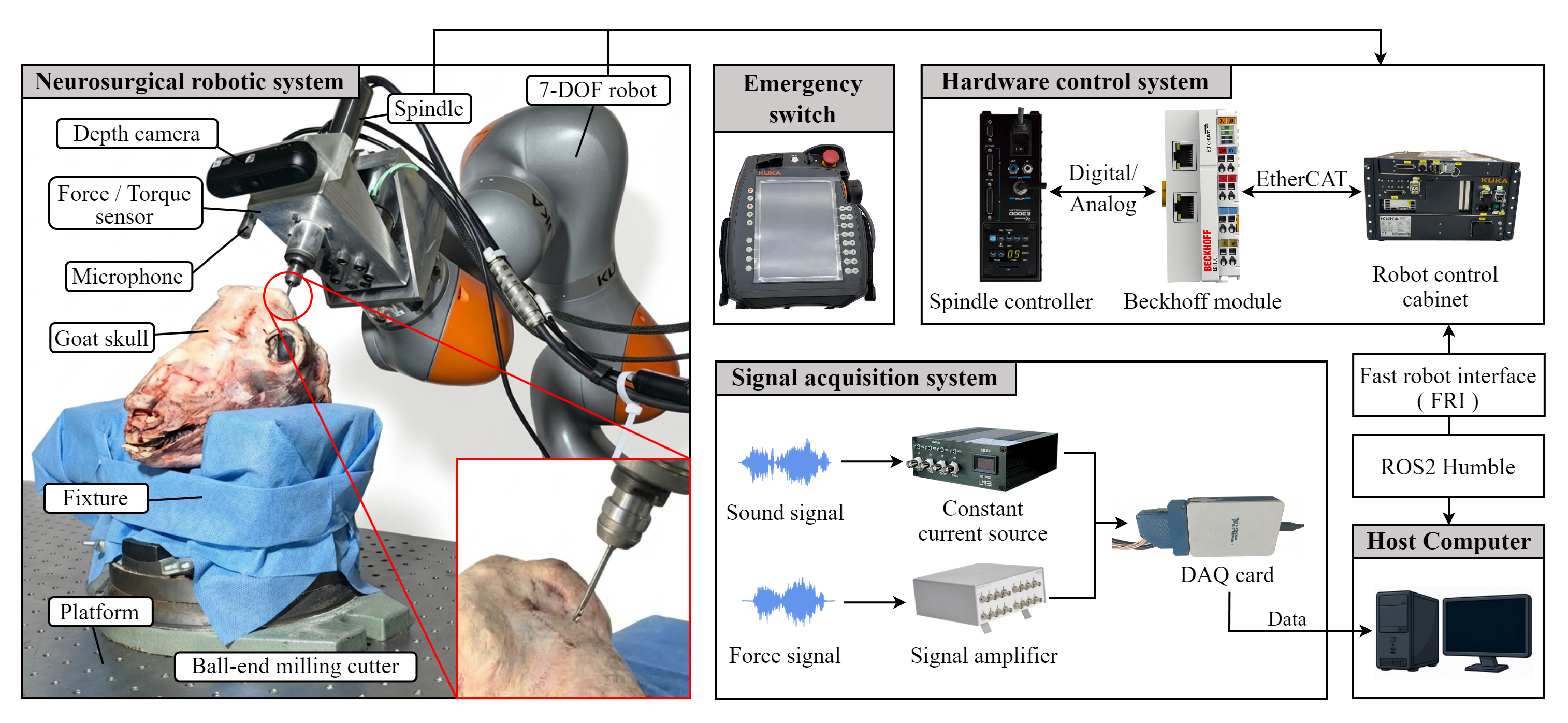}
    \caption{Experimental platform for robotic automatic bone removal.}
    \label{fig:platform}
\end{figure*}

\section{Experimental Skull-Milling System}

An autonomous robotic craniotomy system was developed to validate the proposed framework (Fig.~\ref{fig:platform}). The system comprises a 7-DOF robotic arm (KUKA LBR iiwa 14 R820), a milling module, a signal acquisition module, and a vision-guidance module. The milling module includes a high-speed spindle (Nakanishi), a controller, and a \SI{4}{mm} ball-end cutter. The spindle controller communicates with a Beckhoff system via analog/digital signals, which interfaces with the robot via EtherCAT for real-time spindle-speed regulation. The signal acquisition module integrates a microphone (BSWA MPA201) and a 6-DOF force/torque sensor (Kistler 9306A), with signals conditioned and acquired via a DAQ card (NI-9220) for multimodal monitoring using the CMA-TCN-Transformer with ABF. The vision module employs a depth camera (Orbbec Gemini 305) for point cloud acquisition and registration. The system runs on Ubuntu 22.04 with ROS~2 Humble, achieving a control cycle of \SI{10}{ms} (\SI{100}{Hz}). An emergency stop mechanism is provided via both software and hardware interfaces to ensure safety.

\section{Milling Control Method}
\subsection{Dual-Contour-Fusion-Based Adaptive Spiral Trajectory Planning Method}

\begin{figure}[htbp]
    \centering
    \includegraphics[width=0.9\columnwidth]{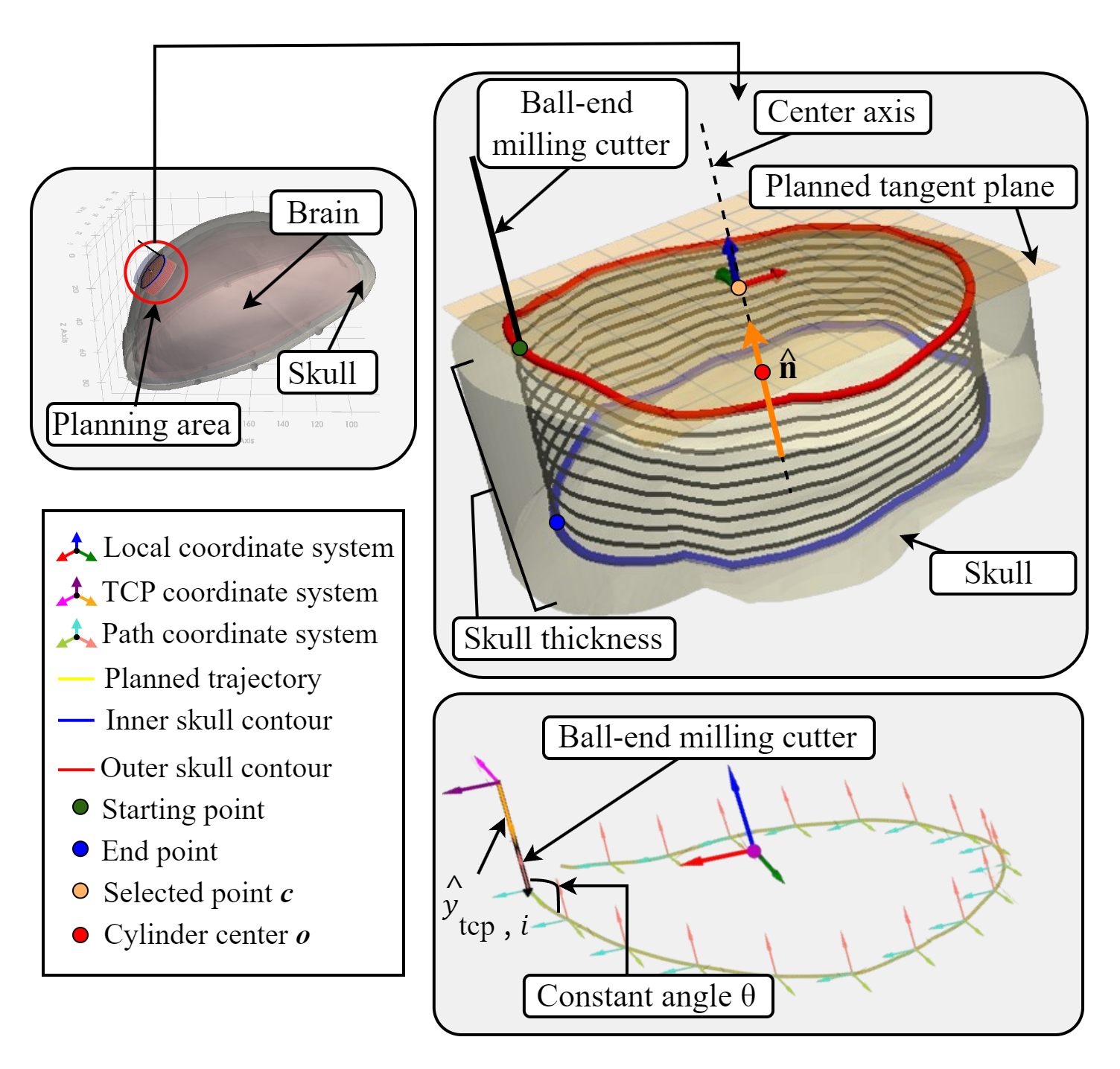}
    \caption{Adaptive spiral path generation method considering the morphology of the outer and inner skull contours.}
    \label{fig:system}
    \vspace{-4mm} 
\end{figure}

In the preoperative stage, a dual-contour-fusion-based adaptive spiral trajectory planning method is developed for robotic craniotomy under complex skull geometry, as illustrated in Fig.~\ref{fig:system}. Unlike conventional preplanned trajectories that require relatively complex preprocessing, the proposed method exploits the outer and inner skull contours to progressively allocate cutting depth, thereby rapidly generating a main trajectory \(\Tmain\) for bulk bone removal, an auxiliary trajectory \(\Taux\) for residual bone removal, and robot-friendly tool posture.

First, a skull mesh model \(\mathcal{M}\) is reconstructed from preoperative CT data. To simplify geometric modeling, the bone-removal volume \(\mathcal{V}\) is approximated as a cylinder extending along a prescribed central axis with unit normal \(\hat{\mathbf{n}}\). To avoid singularities, a reference vector is introduced, and a local right-handed coordinate system \(\{\mathbf{u}, \mathbf{v}, \hat{\mathbf{n}}\}\) is formed via Gram-Schmidt orthogonalization \cite{21}. Let the skull thickness be \(d\), the cylinder radius be \(R\), and its geometric center be \(\mathbf{o}\). The volume is defined as:
\begin{equation}
 \mathcal{V} = \left\{ \mathbf{x} \in \mathbb{R}^3 \,\middle|\,
\left\| \mathbf{r}_i - (\mathbf{r}_i \cdot \hat{\mathbf{n}})\hat{\mathbf{n}} \right\| \le R,\;
\left| \mathbf{r}_i \cdot \hat{\mathbf{n}} \right| \le \frac{d}{2}
\right\}   
\end{equation}
where \(\mathbf{r}_i = \mathbf{x}_i - \mathbf{o}\) denotes the radial vector. A boolean intersection \(\mathcal{I} = \mathcal{M} \cap \mathcal{V}\) is performed to extract the closed contours \(\mathcal{C}_{\text{outer}}\) and \(\mathcal{C}_{\text{inner}}\), which are subsequently mapped into the local cylindrical coordinates \((\phi,\rho,z)\).

To avoid tool overload, the required number of spiral turns \(N_{\text{turns}}\) is determined by the maximum distance \(\Delta H\) between \(\mathcal{C}_{\text{outer}}\) and \(\mathcal{C}_{\text{inner}}\) divided by the allowable cutting depth per turn \(P\). The total rotation angle \(\Theta_{\text{total}} = 2\pi N_{\text{turns}}\) is then uniformly discretized into an angle sequence \(\{\phi_k\}\). A periodic phase mapping \(\tilde{\phi}_k\) is introduced to align the mapped phase within a standard \(2\pi\) period. 

Because spiral trajectories inherently leave geometric residues near the inner skull surface \cite{22}, an auxiliary trajectory \(\Taux\) is explicitly appended using the contour \(\mathcal{C}_{\text{inner}}\). Since the shapes of \(\mathcal{C}_{\text{outer}}\) and \(\mathcal{C}_{\text{inner}}\) are generally inconsistent, a global piecewise adaptive blending function \(\alpha(\phi)\) is introduced to smoothly distribute the cutting depth from the outer contour to the inner contour:
\begin{equation}
\alpha(\phi) =
\begin{cases}
\min\left( \dfrac{\phi - \phi_{\text{start}}}{\Theta_{\text{total}}}, 1 \right), & \phi \in [\phi_{\text{start}}, \phi_{\text{end}}] \\
1, & \phi \in (\phi_{\text{end}}, \phi_{\text{end}} + 2\pi]
\end{cases}
\end{equation}

Accordingly, the final Cartesian trajectory point \(\mathbf{x}(\phi)\) is synthesized via linear interpolation weighted by \(\alpha(\phi)\):
\begin{equation}
  \left\{
\begin{aligned}
\rho(\phi) &= [1-\alpha(\phi)] \rho_{\text{outer}}(\tilde{\phi}) + \alpha(\phi) \rho_{\text{inner}}(\tilde{\phi}) \\
z(\phi) &= [1-\alpha(\phi)] z_{\text{outer}}(\tilde{\phi}) + \alpha(\phi) z_{\text{inner}}(\tilde{\phi}) \\
\mathbf{x}(\phi) &= \mathbf{o} + \rho(\phi)\bigl(\cos\phi\,\mathbf{u} + \sin\phi\,\mathbf{v}\bigr) + z(\phi)\hat{\mathbf{n}}
\end{aligned}
\right.  
\end{equation}

Finally, to improve chip evacuation, the tool axis (the \(Y\)-axis of the TCP frame) is assigned an inclination angle \(\theta\) relative to the local path tangent \cite{23}. To avoid robot kinematic singularities, the global directional preference of the TCP frame is maintained throughout the path. The local \(X\)-axis is orthogonalized against the updated \(Y\)-axis using an initial reference axis \(\hat{\mathbf{x}}_{\text{init}}\), and the complementary \(Z\)-axis is determined via the right-hand rule. This yields the corresponding quaternion sequence \(\{q_i\}\), which is further processed through sign correction, 1D Gaussian filtering, and spherical linear interpolation to ensure continuity. The complete trajectory set is thus generated as \(\mathcal{T}=\{(\mathbf{x}_k,q_k)\}\).

\begin{figure*}[!t]
    \centering
    \includegraphics[width=\textwidth]{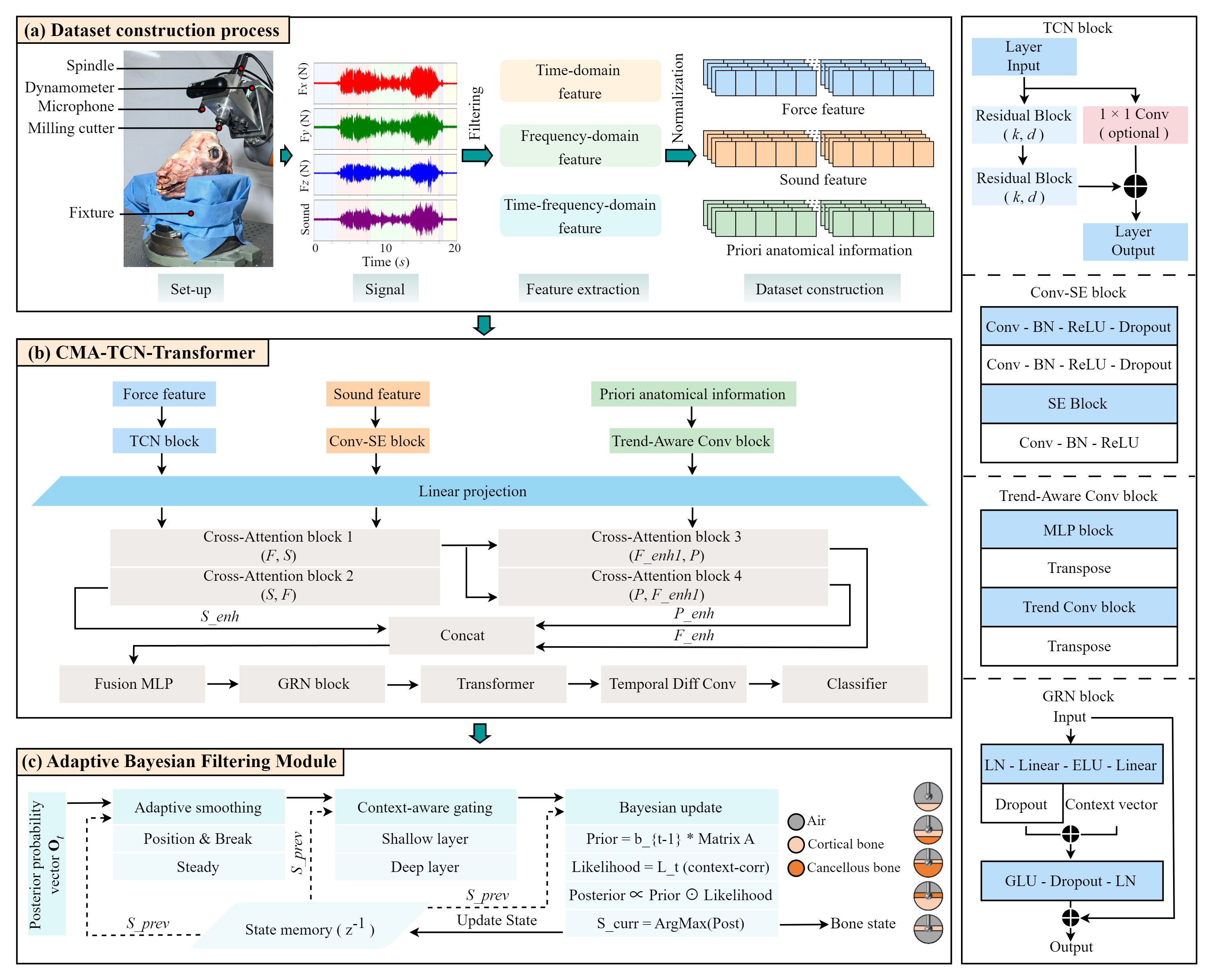}
    \caption{Overall architecture of the CMA-TCN-Transformer network. (a) Data preprocessing module. (b) CMA-TCN-Transformer module. (c) Adaptive Bayesian filtering module.}
    \label{fig:CMA_arch}
\end{figure*}

\subsection{Intraoperative Bone-Removal State Monitoring Based on CM A-TCN-Transformer with ABF}

\begin{figure*}[t]
    \centering
    \includegraphics[width=\textwidth]{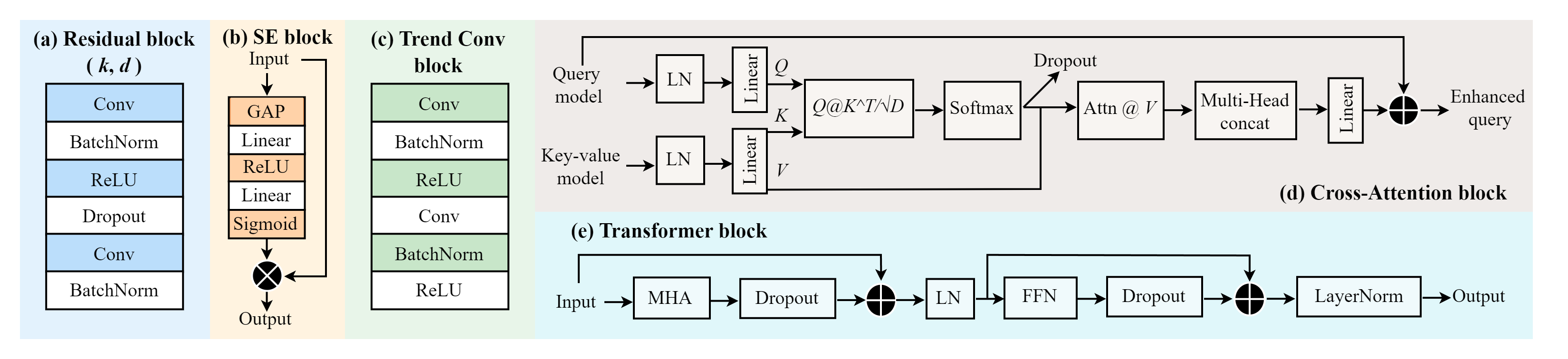}
    \vspace{-6mm} 
    \caption{Detailed structures of key blocks in the CMA-TCN-Transformer. (a) Residual block. (b) SE block. (c) Trend Conv block. (d) Cross-attention block. (e) Transformer block.}
    \label{fig:CMA_blocks}
    \vspace{-4mm}
\end{figure*}

During intraoperative milling, the robot removes bone along the predefined main trajectory \(\Tmain\). However, due to heterogeneous bone density, weak robot stiffness, and intraoperative disturbances, single-signal monitoring is insufficient for robust bone-layer state recognition. Inspired by how experienced surgeons jointly interpret cutting force, sound variation, and feed progress, a human-inspired CMA-TCN-Transformer with ABF is developed (Fig.~\ref{fig:CMA_arch}). By coupling force and sound signals, the proposed method enables robust online identification of bone-layer states and low-latency breakthrough monitoring.

In the data preprocessing stage (Fig.~\ref{fig:CMA_arch}(a)), force and sound signals are first filtered and segmented into aligned temporal windows. In the feature extraction stage (Fig.~\ref{fig:CMA_arch}(b)), the model separately extracts time-frequency domain features from the force and sound signals, and captures \textit{a priori} anatomical information (i.e., the inherent cortical-cancellous-cortical bone structure). 

These representations are fed into a three-branch parallel architecture. For force signals, a temporal convolutional network (TCN) composed of cascaded dilated convolutions and residual connections (Fig.~\ref{fig:CMA_blocks}(a)) is introduced to enlarge the temporal receptive field without significantly increasing computational complexity, yielding the force feature \(\mathbf{H}_F\). For sound signals, multilayer 1D convolutions extract local spectral patterns, followed by a squeeze and excitation (SE) module (Fig.~\ref{fig:CMA_blocks}(b)) to suppress machining noise via channel-wise dynamic recalibration, producing the enhanced sound feature \(\mathbf{H}_S\). The \textit{priori} anatomical information branch uses a multilayer perceptron (MLP) together with a trend-aware convolution module (Fig.~\ref{fig:CMA_blocks}(c)) to generate the position feature \(\mathbf{H}_P\), injecting bone-removal progress information to alleviate classification ambiguity.

To achieve effective multimodal fusion, a two-stage cross-modal attention block (CMA) (Fig.~\ref{fig:CMA_blocks}(d)) sequentially performs bidirectional force--sound and force--position coupling, which is consistent with prior studies on surgical skill assessment demonstrating that multimodal fusion can significantly improve robustness and accuracy in complex manipulation recognition tasks \cite{24}. The concatenated fused feature \(\mathbf{H}_{\mathrm{fused}}\) is refined by a gated residual network (GRN), which acts as an adaptive filter to regulate effective information flow. The refined feature sequence is then input into a Transformer (Fig.~\ref{fig:CMA_blocks}(e)) for temporal modeling. Although the Transformer is effective in capturing long-range temporal dependencies, its self-attention mechanism may over-smooth sharp local variations associated with breakthrough. To alleviate this issue, a temporal-difference convolution module is cascaded afterward as a high-pass enhancement filter to explicitly amplify local gradient changes:
\begin{equation}
        \tilde{\mathbf{H}}_{\mathrm{enc}} = \mathbf{H}_{\mathrm{enc}} + \mathrm{ReLU}\!\left( \mathrm{BN}\!\left( \mathrm{Conv}_{k_{td}}\!\left(\mathbf{H}_{\mathrm{enc}}^\top\right) \right) \right)^\top
\end{equation}

where \(\mathbf{H}_{\mathrm{enc}}\) is the Transformer output, and \(\tilde{\mathbf{H}}_{\mathrm{enc}}\) is the final feature after differential enhancement. The feature at the final time step is processed by a perceptron and a Softmax layer, yielding the posterior probability vector \(\mathbf{O}_t\) over the \(C\) bone-layer states.

Although the CMA-TCN-Transformer provides accurate instantaneous estimates, heterogeneous bone density and local disturbances may still induce transient predictive fluctuations, potentially leading to premature breakthrough predictions \cite{25,26}. To suppress such fluctuations, ABF is introduced (Fig.~\ref{fig:CMA_arch}(c)). ABF combines a Markovian state-transition model with surgical-progress-aware likelihood modulation to smooth the raw probability sequence\cite{27}. Specifically, a context-aware dynamic likelihood vector \(\mathbf{L}_t\) is constructed to explicitly penalize implausible cross-layer state transitions while adaptively enhancing sensitivity to weak breakthrough signals. The state confidence \(\mathbf{b}_t\) is recursively updated as:
\begin{equation}
        \begin{aligned}
        \mathbf{b}_{\mathrm{prior}} &= \mathbf{b}_{t-1} \cdot \mathbf{A}, \quad
        \mathbf{b}_{\mathrm{post}} = \mathbf{b}_{\mathrm{prior}} \odot (\mathbf{L}_t + \varepsilon) \\
        \mathbf{b}_t &= \mathbf{b}_{\mathrm{post}} \,/\, \|\mathbf{b}_{\mathrm{post}}\|_1
    \end{aligned}
\end{equation}

where \(\mathbf{A}\) is the state-transition matrix encoding the hard constraint of unidirectional breakthrough progression, and \(\varepsilon\) prevents numerical underflow. To ensure surgical safety, once the optimal posterior estimate \(s_t = \arg\max(\mathbf{b}_t)\) first identifies the breakthrough state, the confidence vector is permanently locked. The complete estimation procedure is summarized in Algorithm~\ref{alg:bayesian_estimation}.

\begin{algorithm}[!t]
\caption{Adaptive Bayesian State Estimation}
\label{alg:bayesian_estimation}
\small
\begin{algorithmic}[1]
\REQUIRE $\mathbf{O}_t \in [0,1]^C$; Prior $\mathbf{b}_{t-1}$, obs $\bar{\mathbf{O}}_{t-1}$; Trans $\mathbf{A}$; Params $\tau, \alpha_{\mathrm{st}}, \alpha_{\mathrm{tr}}, \lambda_{\mathrm{pen}}, \lambda_{\mathrm{bst}}, \mathbf{w}, \varepsilon$; \texttt{locked}.
\ENSURE Estimate $s_t$; Posterior $\mathbf{b}_t$; Smoothed obs $\bar{\mathbf{O}}_t$.

\IF{\texttt{locked}}
    \STATE \textbf{return} $S_{\mathrm{Breakthrough}},\;\mathbf{e}_5,\;\bar{\mathbf{O}}_{t-1}$
\ENDIF
\STATE $\hat{s}_{\mathrm{prev}} \leftarrow \arg\max(\mathbf{b}_{t-1})$
\STATE $\alpha_t \leftarrow (\hat{s}_{\mathrm{prev}} {=} S_{\mathrm{Cort2}} \land \mathbf{O}_t[S_{\mathrm{Breakthrough}}] {>} \tau) \;?\; \alpha_{\mathrm{tr}} : \alpha_{\mathrm{st}}$
\STATE $\bar{\mathbf{O}}_t \leftarrow \mathbf{L}_t \leftarrow \alpha_t \mathbf{O}_t + (1-\alpha_t)\bar{\mathbf{O}}_{t-1}$

\STATE \textit{\% Context-aware gating mechanism}
\IF{$\hat{s}_{\mathrm{prev}} \in \{S_{\mathrm{Air}}, S_{\mathrm{Cort1}}, S_{\mathrm{Canc}}\}$}
    \STATE $\delta \leftarrow \mathbf{L}_t[S_{\mathrm{Cort2}}] \cdot (1 - \lambda_{\mathrm{pen}})$
    \STATE $\mathbf{L}_t[\{S_{\mathrm{Cort1}}, S_{\mathrm{Canc}}\}] \mathrel{{+}{=}} \mathbf{w} \delta$
    \STATE $\mathbf{L}_t[\{S_{\mathrm{Cort2}}, S_{\mathrm{Break}}\}] \mathrel{{*}{=}} \lambda_{\mathrm{pen}}$
\ELSIF{$\hat{s}_{\mathrm{prev}} = S_{\mathrm{Cort2}}$}
    \STATE $\mathbf{L}_t[S_{\mathrm{Break}}] \mathrel{{*}{=}} \lambda_{\mathrm{bst}}$ \COMMENT{Sensitivity boost}
\ENDIF

\STATE \textit{\% Bayesian update \& lock trigger}
\STATE $\mathbf{b}_t \leftarrow \left((\mathbf{b}_{t-1} \mathbf{A}) \odot (\mathbf{L}_t + \varepsilon)\right) / \| \cdot \|_1$
\STATE $s_t \leftarrow \arg\max(\mathbf{b}_t)$
\IF{$s_t = S_{\mathrm{Breakthrough}}$}
    \STATE $\mathbf{b}_t \leftarrow \mathbf{e}_5;\quad \texttt{locked} \leftarrow \texttt{True}$
\ENDIF
\STATE \textbf{return} $s_t,\;\mathbf{b}_t,\;\bar{\mathbf{O}}_t$
\end{algorithmic}
\end{algorithm}

\begin{figure}[htbp]
    \centering
    \includegraphics[width=\columnwidth]{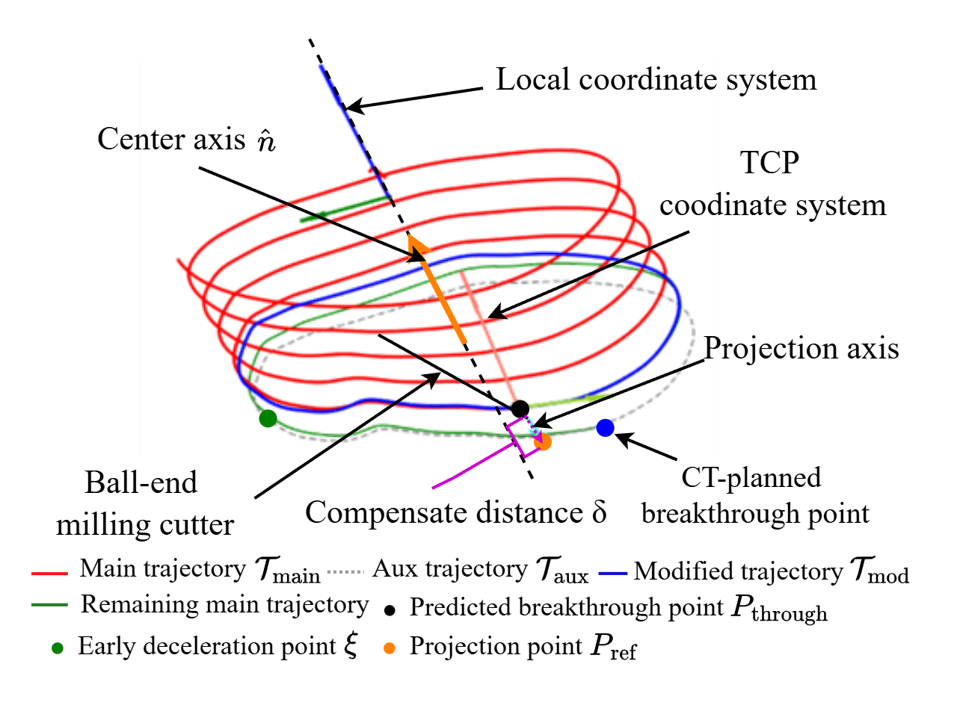}
    \caption{Breakthrough-triggered trajectory adjustment strategy with distance compensation along the projection axis}
    \label{fig:trajectory_adj}
\end{figure}

\subsection{Breakthrough-Triggered \textit{in-situ} Projection-Based Trajectory Adjustment Strategy}

After the breakthrough state is identified during intraoperative bone removal, directly invoking the predefined auxiliary trajectory \(\Taux\) may introduce depth mismatch and increase the risk of dural injury. To address this issue, a breakthrough-triggered \textit{in-situ} projection-based trajectory adjustment strategy is developed for safe residual bone removal, as shown in fig.~\ref{fig:trajectory_adj}. The key idea is to use the predicted breakthrough point as an intraoperative geometric reference, project it onto the auxiliary trajectory along the local normal direction to obtain a reference point, and then compute the axial compensation vector between the two points to uniformly translate the auxiliary trajectory before residual bone removal.

Before automatic bone removal begins, the preplanned trajectories \(\Tmain\) and \(\Taux\) are transformed into the robot base frame through point-cloud registration and eye-in-hand calibration \cite{28}. The robot then starts automatic bone removal along the main trajectory \(\Tmain\). Let the execution progress along \(\Tmain\) be denoted by \(\xi \in [0,1]\). Because local stress release near breakthrough may induce end-effector overshoot during the final stage of milling, a variable milling feed-speed strategy is adopted. During the normal bone-removal stage (\(\xi < 0.9\)), the robot feeds at the prescribed speed \(v_{\text{set}}\). When the tool approaches the estimated inner cortical layer (\(\xi \geq 0.9\)), the feed speed is smoothly reduced to the safety threshold \(v_{\text{safe}}=\SI{1}{mm/s}\). Throughout the entire bone-removal process, the CMA-TCN-Transformer with ABF continuously performs online monitoring of the current bone-layer state. Once the breakthrough state is identified, the robot controller immediately interrupts the execution of \(\Tmain\) and latches the actual breakthrough pose \(\mathbf{p}_{\text{breakthrough}}\).

Due to registration errors, heterogeneous bone density, and intraoperative skull displacement, the actual breakthrough point \(\mathbf{p}_{\text{breakthrough}}\) generally deviates from the theoretical breakthrough point along the local depth direction \(\hat{\mathbf{n}}\)\cite{29,30}. Therefore, directly invoking the predefined auxiliary trajectory \(\Taux\) for residual bone removal may increase the risk of dural injury. To compensate for this deviation, an \textit{in-situ} projection-based adjustment strategy is introduced. Specifically, \(\mathbf{p}_{\text{breakthrough}}\) is projected onto \(\Taux\) along the local normal direction \(\hat{\mathbf{n}}\) within the local coordinate system \(\{\mathbf{u}, \mathbf{v}, \hat{\mathbf{n}}\}\), yielding the reference point \(\mathbf{p}_{\text{ref}}\). The axial compensation vector \(\boldsymbol{\delta}\) between the two points is then calculated as
\begin{equation}
    \boldsymbol{\delta}
    =
    \left[
    \left(\mathbf{p}_{\text{breakthrough}}-\mathbf{p}_{\text{ref}}\right)\cdot\hat{\mathbf{n}}
    \right]\hat{\mathbf{n}}
\end{equation}

The computed \(\boldsymbol{\delta}\) is uniformly applied to all discrete waypoints \(\mathbf{x}_i\) on \(\Taux\), i.e., as a translation along the \(\hat{\mathbf{n}}\) direction, thereby generating the modified auxiliary trajectory \(\Tmod\). The robot then follows \(\Tmod\) to complete safe residual bone removal and achieve automated bone-flap separation. The complete closed-loop execution procedure is summarized in Algorithm~\ref{alg:milling_control}.

\begin{algorithm}[!t]
\caption{Breakthrough-Triggered Trajectory Adjustment for Residual Bone Removal}
\label{alg:milling_control}
\small
\begin{algorithmic}[2]
\REQUIRE Main trajectory $\Tmain$, auxiliary trajectory $\Taux$; registration matrix $\mathbf{T}_{\mathrm{reg}}$; velocity thresholds $v_{\mathrm{set}}, v_{\mathrm{safe}}$; safety progress threshold $\xi_{\mathrm{safe}}$; local normal vector $\hat{\mathbf{n}}$.
\ENSURE Safe completion of residual bone removal without dural injury.

\STATE \textit{\% Initialization: map trajectories into the robot base frame}
\STATE $\Tmain' \leftarrow \mathbf{T}_{\mathrm{reg}} \cdot \Tmain$
\STATE $\Taux' \leftarrow \mathbf{T}_{\mathrm{reg}} \cdot \Taux$
\STATE $s_t \leftarrow S_{\mathrm{Normal}}$

\STATE \textit{\% Stage 1: main-trajectory execution and breakthrough monitoring}
\FOR{$k = 1$ \textbf{to} $|\Tmain'|$}
    \STATE $\xi \leftarrow k / |\Tmain'|$
    \IF{$\xi < \xi_{\mathrm{safe}}$}
        \STATE $v_{\mathrm{cmd}} \leftarrow v_{\mathrm{set}}$
    \ELSE
        \STATE $v_{\mathrm{cmd}} \leftarrow v_{\mathrm{safe}}$
    \ENDIF
    \STATE $\mathbf{X}_{\mathrm{sens}} \leftarrow \text{ReadMultimodalSensors}()$
    \STATE $s_t \leftarrow \text{Alg.~\ref{alg:bayesian_estimation}}\big(\text{CMA-TCN-Transformer}(\mathbf{X}_{\mathrm{sens}})\big)$
    \IF{$s_t = S_{\mathrm{Breakthrough}}$}
        \STATE $\mathbf{p}_{\mathrm{breakthrough}} \leftarrow \text{GetEndEffectorPose}()$
        \STATE \textbf{break}
    \ENDIF
    \STATE $\text{ServoControlTo}(\Tmain'[k], v_{\mathrm{cmd}})$
\ENDFOR

\STATE \textit{\% Stage 2: \textit{in-situ} projection-based auxiliary-trajectory adjustment}
\IF{$s_t = S_{\mathrm{Breakthrough}}$}
    \STATE $\mathbf{p}_{\mathrm{ref}} \leftarrow \text{Proj}_{\hat{\mathbf{n}}}(\mathbf{p}_{\mathrm{breakthrough}}, \Taux')$
    \STATE $\boldsymbol{\delta} \leftarrow \left[(\mathbf{p}_{\mathrm{breakthrough}} - \mathbf{p}_{\mathrm{ref}})\cdot\hat{\mathbf{n}}\right]\hat{\mathbf{n}}$
    \STATE $\Tmod \leftarrow \{\mathbf{x}_i + \boldsymbol{\delta}\mid \mathbf{x}_i \in \Taux'\}$
    \STATE $\text{ExecuteCompensatedPath}(\Tmod, v_{\mathrm{safe}})$
\ENDIF

\STATE $\text{SafeRetract}()$
\end{algorithmic}
\end{algorithm}

\begin{figure}[!t] 
    \centering 
    \includegraphics[width=\columnwidth]{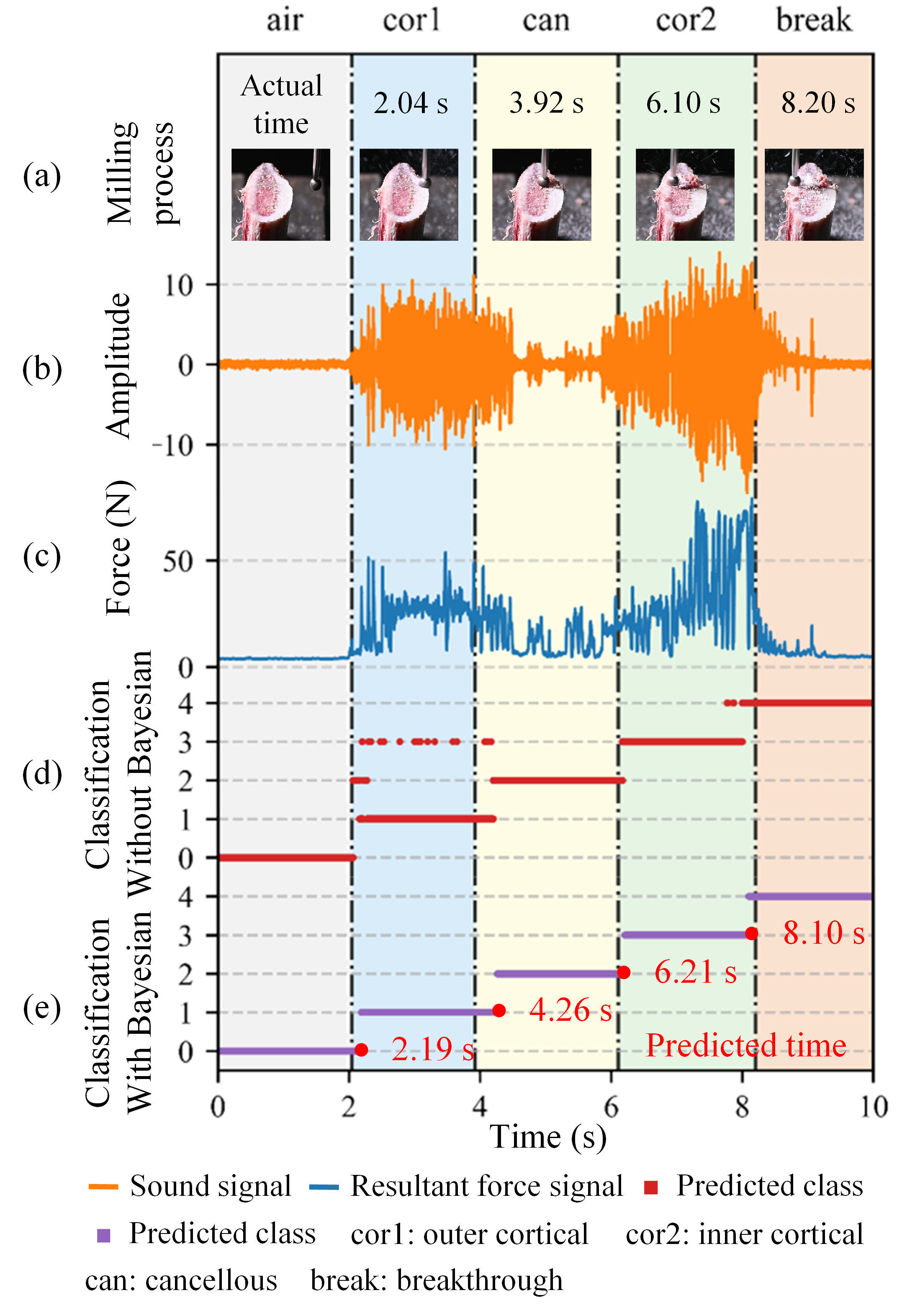} 
    \caption{Bone layer state prediction for robotic rib milling with online deployed CMA‑TCN‑Transformer . (a) Milling process. (b) Sound signal . (c) Force signal. (d) Prediction results without the ABF. (e) Prediction results with the ABF. The labels 0, 1, 2, 3, and 4 denote air cut, outer cortical bone, cancellous bone, inner cortical bone, and breakthrough, respectively.} 
    \label{fig:online_prediction} 
    \vspace{-3.5mm}
\end{figure}

\section{Analysis of Experimental Results}
The effectiveness of the proposed framework is validated through two groups of experiments. Bovine rib milling experiments are first conducted to evaluate the performance of the proposed ABF-coupled CMA-TCN-Transformer in terms of bone-layer monitoring accuracy and breakthrough response latency. Furthermore, autonomous robotic craniotomy experiments on \textit{ex vivo} goat skulls are performed to verify the safety and effectiveness of the overall framework under complex skull-surface geometry.
\subsection{Bovine rib milling experiment}
Bovine ribs were selected as the experimental model because they provide a practical and repeatable ex vivo specimen with a distinct cortical-cancellous-cortical structure, which is suitable for evaluating breakthrough-related state transitions during bone milling\cite{31}.
Fig.~\ref{fig:online_prediction}(a) presents a complete rib-milling experiment. A DSLR camera synchronously recorded the entire cutting process to provide reliable ground-truth labels, which were time-aligned with the multi-source sensor data. During milling, the cutter remained perpendicular to the pre-flattened bone surface. Fig.~\ref{fig:online_prediction}(b) and (c) show the recorded sound and force signals, respectively. Based on these synchronized data, the online prediction results are shown in Fig.~\ref{fig:online_prediction}(d) and (e). Without the ABF module, the network still produces occasional false breakthrough predictions in the inner cortical bone region due to fluctuations caused by heterogeneous bone density, which may lead to premature trajectory switching and incomplete bone removal. After introducing the ABF module, the CMA-TCN-Transformer effectively suppresses such transient interference and yields continuous and robust state monitoring results.

\begin{figure}[!t] 
    \centering 
    \includegraphics[width=\columnwidth]{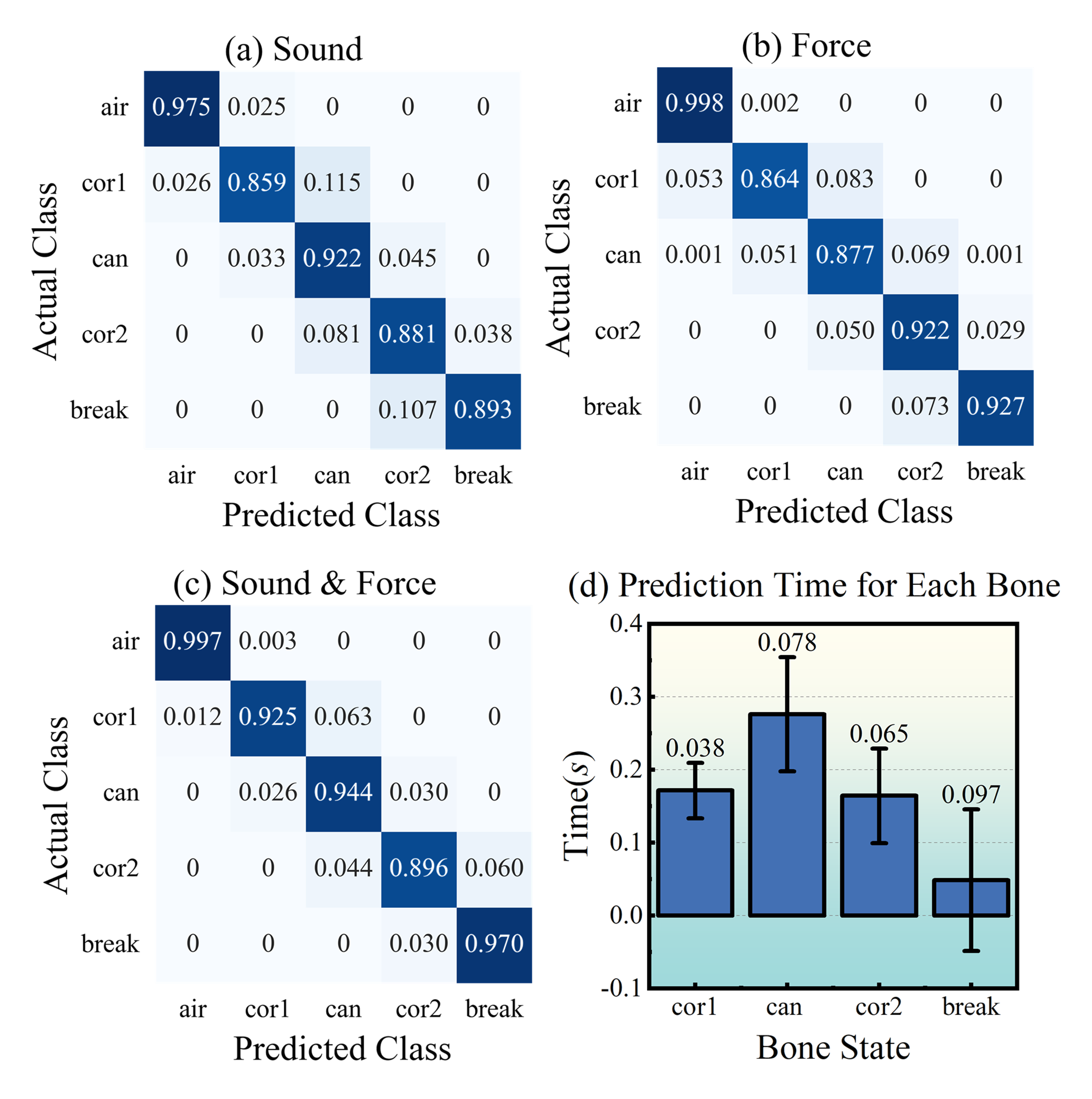} 
    \caption{Bone layer state prediction results and temporal distributions of the CMA-TCN-Transformer under different input signals. (a) Without force. (b) Without sound. (c) All signal. (d) Temporal distribution of bone layer state prediction with all signal inputs. } 
    \label{fig:confusion_matrices} 
\end{figure}

Fig.~\ref{fig:confusion_matrices} further compares the performance of different input modalities. Relying solely on sound signals, the model achieves a breakthrough prediction accuracy of 89.3\% (Fig.~\ref{fig:confusion_matrices}(a)). With the force signal alone, the accuracy increases to 92.7\% (Fig.~\ref{fig:confusion_matrices}(b)), indicating stronger robustness, although false positives still occur during breakthrough detection. After fusing both force and sound signals, the breakthrough prediction accuracy further improves to 97\% (Fig.~\ref{fig:confusion_matrices}(c)), which demonstrates the benefit of multimodal information fusion.

As shown in Fig.~\ref{fig:confusion_matrices}(d), statistics from 16 experiments indicate that the detection delays for entering the outer cortical bone, cancellous bone, and inner cortical bone are \(0.17 \pm 0.038\) s, \(0.27 \pm 0.078\) s, and \(0.16 \pm 0.065\) s, respectively, while the breakthrough recognition delay is only \(0.048 \pm 0.097\) s. The slightly longer delay in cancellous bone is mainly attributable to its gradual mechanical transition from the adjacent cortical bone. The relatively large standard deviation in breakthrough detection may result from the network's sensitivity to subtle precursor features before visible breakthrough occurs.

\begin{figure}[!t] 
    \centering 
    \includegraphics[width=\columnwidth]{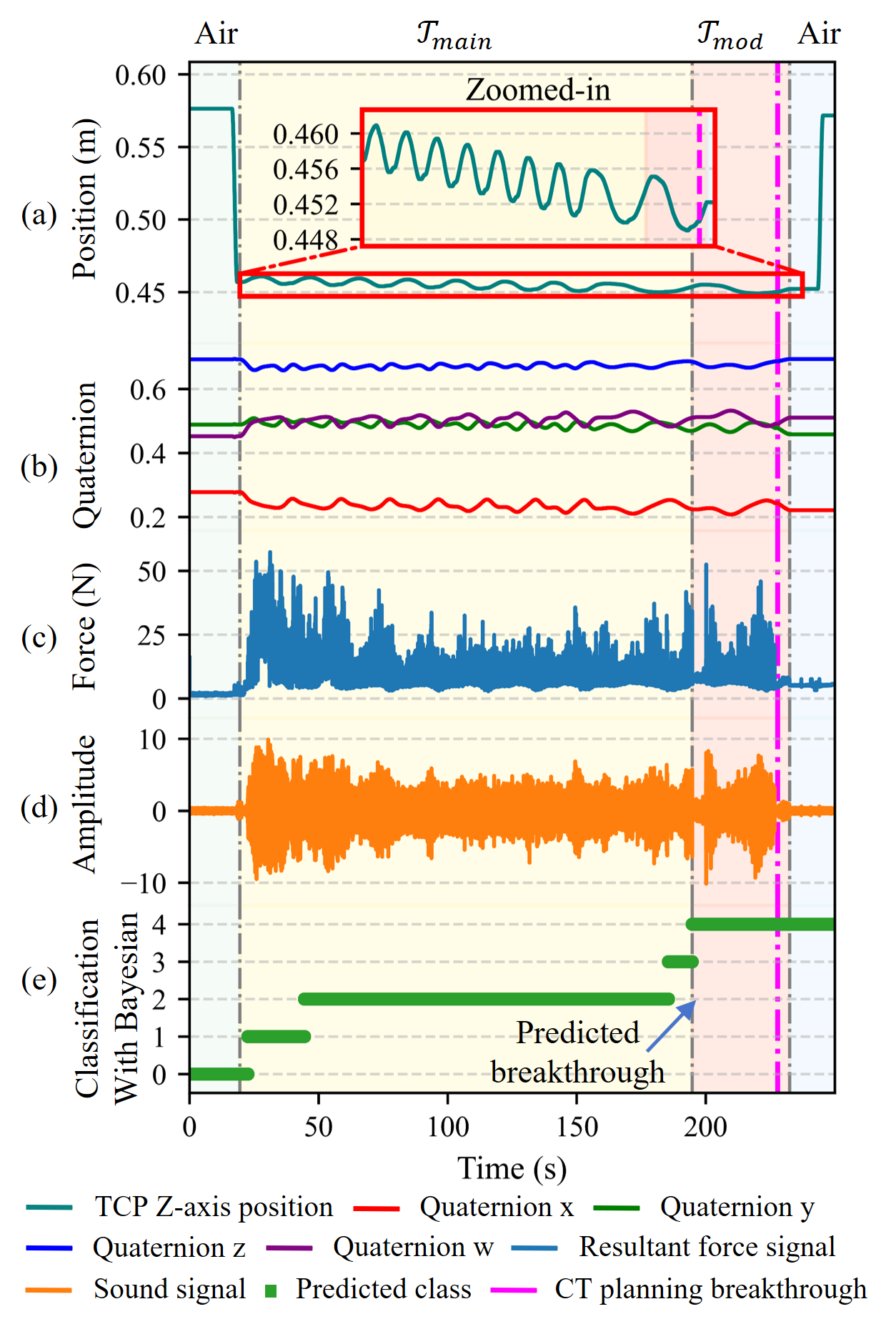} 
    \caption{Experimental results of the robotic skull milling process. (a) Variation of the TCP Z-coordinate. (b) Variation of the TCP Orientation. (c) Force signal. (d) Sound signal. (e) Prediction results of the CMA-TCN-Transformer with ABF . The labels 0, 1, 2, 3, and 4 denote air cut, outer cortical bone, cancellous bone, inner cortical bone, and breakthrough, respectively.} 
    \label{skull milling} 
\end{figure}

\begin{figure*}[!t] 
    \centering 
    \includegraphics[width=\textwidth]{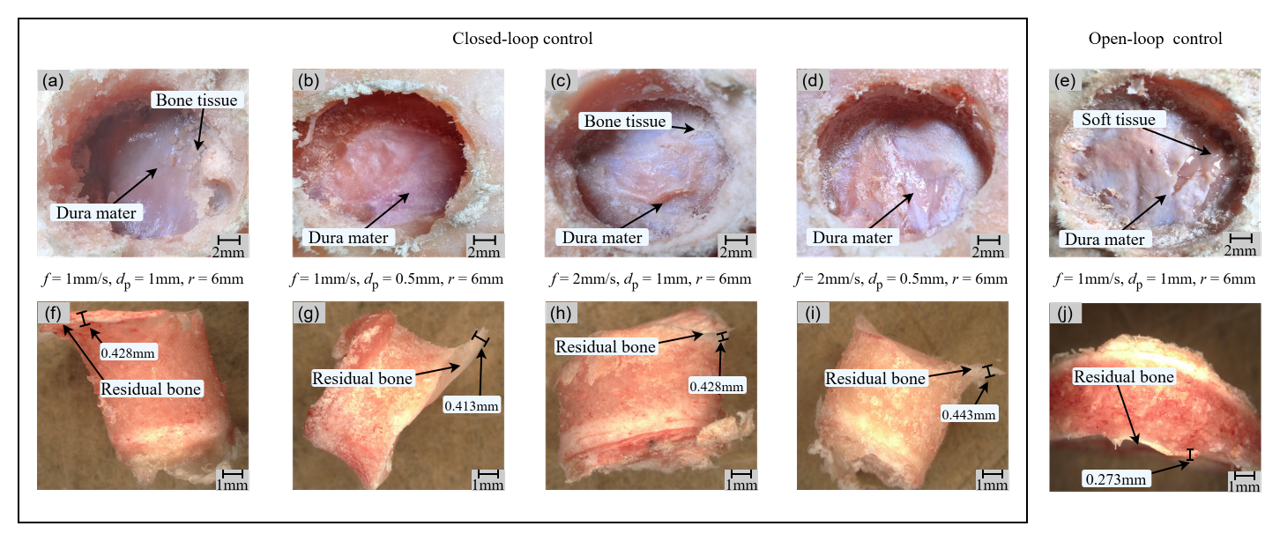} 
    \caption{Photographs of burr holes and corresponding extracted bone flaps after automated robotic bone removal under different parameters. (a)--(e) Burr holes. (f)--(j) Extracted bone flaps.} 
    \label{fig:bone removal} 
\end{figure*}

\subsection{Autonomous robotic craniotomy experiments on \textit{ex vivo} goat skulls}

To further verify the safety and effectiveness of the proposed autonomous robotic craniotomy framework under complex skull geometries, experiments were conducted on \textit{ex vivo} goat skulls. Goat skulls were selected due to their thickness distribution (4–20 mm), comprehensively encompassing the typical human skull thickness range of 4–14 mm \cite{32,33}. We performed five bone-flap removal experiments under varying feed rates, cutting depths, and craniotomy radii. These included four closed-loop experiments (utilizing the proposed trajectory adjustment strategy), and one open-loop baseline experiment (without this strategy). Surgical success was strictly defined by two criteria: the smooth detachment of the bone flap using forceps post-milling, and the absolute intactness of the underlying dura mater.

As shown in Fig.~\ref{skull milling}(a), the $Z$-axis coordinate of the end-effector exhibits periodic variations with an increasing period, reflecting the effect of the adaptive fusion function $\alpha(\phi)$, which enables smooth transition from the outer to the inner skull surface while maintaining cutting depth. At the predicted breakthrough point, the robot transitions from $\Taux$ to $\Tmod$ without noticeable overshoot and continues smoothly from $\Tmain$ to $\Tmod$. The quaternion curves in Fig.~\ref{skull milling}(b) show only small fluctuations, indicating stable tool orientation under the proposed optimization strategy.
The force and sound signals in Fig.~\ref{skull milling}(c) and (d) exhibit significant fluctuations, suggesting that simple threshold-based methods are insufficient for reliable state monitoring. As shown in Fig.~\ref{skull milling}(e), the predicted breakthrough occurs approximately one spiral revolution earlier than the CT-based estimate, demonstrating that open-loop execution based solely on CT is prone to over-cutting. This validates the necessity of the proposed \emph{in situ} trajectory compensation strategy for safe operation.

The final flap-removal results are shown in Fig.~\ref{fig:bone removal}. The four closed-loop craniotomy experiments (Fig.~\ref{fig:bone removal}(a)--(d)) were all successfully completed without dural injury, whereas the open-loop experiment (Fig.~\ref{fig:bone removal}(e)) failed because the dura mater was injured. In the closed-loop experiments, the detached bone flaps generally retained either a partial residual bone layer or no obvious residual layer, indicating effective and safe depth regulation near the inner cortical boundary. The residual bone-layer thickness after closed-loop craniotomy was \(0.428 \pm 0.015\) mm. As shown in Fig.~\ref{fig:bone removal}(f)--(i), the proposed framework achieved favorable performance for bone-flap removal with different feed rates, cutting depths, flap sizes, and skull thicknesses, demonstrating the safety, effectiveness, and robustness of the overall framework under complex skull-surface geometry.

\section{Discussion}

\begin{table*}[t]
\begin{center}
\caption{Comparison with representative robot-assisted craniotomy studies.}
\label{tab:comparison}
\footnotesize
\setlength{\tabcolsep}{4pt}
\renewcommand{\arraystretch}{1.15}

\newcolumntype{C}[1]{>{\centering\arraybackslash}m{#1}}

\resizebox{\textwidth}{!}{%
\begin{tabular}{C{1.8cm} C{1.6cm} C{1.2cm} C{1.6cm} C{2.2cm} C{1.9cm} C{2.0cm} C{1.4cm} C{1.0cm} C{1.0cm}}
\toprule

\textbf{Reference} & \textbf{Trajectory planning} & \textbf{Trajectory Type} & \textbf{Monitoring basis} & \textbf{Adaptation to Bone Geometry} & \textbf{Breakthrough accuracy} & \textbf{Breakthrough latency} & \textbf{Residual-bone handling} & \multicolumn{2}{c}{\textbf{Validation platform}} \\
\cmidrule(lr){9-10}
& & & & & & & & \textbf{Tool} & \textbf{Material} \\

\midrule
Popovic et al. \cite{10} 
& CT-based planning 
& Contour 
& NR 
& Outer surface 
& NR 
& NR 
& NR 
& NR 
& NR \\

Liu et al. \cite{34} 
& CT-based planning 
& Contour 
& Force 
& Outer surface 
& NR 
& NR  
& NR 
& Robot 
& Skull model \\

Li et al. \cite{12} 
& CT-based planning 
& Contour 
& NR 
& Outer surface+relative pose 
& NR 
& NR 
& NR 
& NR 
& NR \\

Osa et al. \cite{35} 
& NR 
& Linear 
& Motor current and IMU 
& NR 
& 75\% 
& NR 
& Limited 
& Hand-held 
& Artificial bone \\

Bian et al. \cite{13,14} 
& NR 
& Linear 
& Force 
& Inner surface 
& 93.06\%(drill) 98.61\%(mill) 
& $>$5 s ($<1$ mm overshoot) 
& Limited 
& Robot 
& Animal skull \\

Sun et al. \cite{17} 
& NR 
& Linear 
& Sound 
& NR 
& 99.12\% 
& NR 
& NR 
& Robot 
& Artificial bone \\

\textbf{Proposed} 
& \textbf{CT-based planning} 
& \textbf{Spiral} 
& \textbf{Force+sound fusion} 
& \textbf{Outer/inner surface+relative pose} 
& \textbf{97\%} 
& \textbf{48 $\pm$ 97 ms ($<0.29$ mm overshoot)} 
& \textbf{Well addressed} 
& \textbf{Robot} 
& \textbf{Animal skull} \\

\bottomrule
\end{tabular}%
} 
\end{center}
\end{table*}

As summarized in Table~\ref{tab:comparison}, existing robotic craniotomy methods remain limited in intraoperative perception, geometric adaptability, and residual-bone handling. Preoperative planning-based approaches \cite{10,12,34} rely on CT-derived trajectories and lack real-time state perception, making them unable to compensate for intraoperative physical deviations. Sensor-based methods enable closed-loop control, but still exhibit limitations. Osa et al. \cite{35} used motor current and IMU signals for process monitoring and breakthrough detection, but the overall accuracy was limited by complex bone geometry. Bian et al. \cite{13,14} achieved breakthrough detection during drilling and milling using force sensing, but their method still follows the conventional two-step procedure and may leave residual bone requiring manual removal. Sun et al. \cite{17} employed a CNN--LSTM network based on sound signals and achieved high accuracy, but validation was limited to artificial bone blocks. In contrast, the proposed framework integrates CT-based adaptive dual-contour fusion planning, multimodal intraoperative perception, and breakthrough-triggered trajectory compensation into a unified closed-loop system, enabling safe breakthrough response and automated residual-bone isolation.

The CMA-TCN-Transformer with ABF suppresses false alarms caused by heterogeneous skull density. Despite a small detection delay, the maximum overshoot is limited to \(\SI{0.29}{mm}\) at \(\SI{2}{mm/s}\), remaining below the dura safety margin \cite{36}. \textit{Ex vivo} experiments further show that actual breakthrough may occur earlier than CT-based estimates due to registration error, bone heterogeneity, and skull displacement, highlighting the necessity of \textit{in-situ} trajectory compensation. All closed-loop goat-skull experiments were completed without dural injury, while the open-loop trial failed. The resulting bone flaps showed minimal residual bone, with a thickness of \(0.428 \pm 0.015\) mm, demonstrating the safety and effectiveness of the proposed framework.

Several limitations remain. CT alone cannot reliably identify pathological soft-tissue conditions, and future work should incorporate MRI into multimodal preoperative planning. Moreover, since the current validation was limited to \textit{ex vivo} specimens, future studies in \textit{in vivo} animal models are needed to assess robustness under physiological disturbances such as bleeding, pulsation, and soft-tissue motion.

\section{Conclusion}
This article proposed a human-inspired framework for autonomous robotic craniotomy, integrating dual-contour trajectory planning, multimodal perception, and breakthrough-triggered \textit{in-situ} compensation. The system achieves 97\% breakthrough detection accuracy with a \(0.048\)~s latency. \textit{Ex vivo} experiments demonstrated safe, autonomous bone flap removal under complex geometries with minimal overshoot, avoiding dural injury entirely. This closed-loop approach effectively bridges the gap between preoperative imaging and intraoperative execution.

\end{document}